\documentclass{article}

\usepackage{arxiv}

\usepackage[utf8]{inputenc} % allow utf-8 input
\usepackage[T1]{fontenc}    % use 8-bit T1 fonts
\usepackage{hyperref}       % hyperlinks
\usepackage{url}            % simple URL typesetting
\usepackage{booktabs}       % professional-quality tables
\usepackage{amsfonts}       % blackboard math symbols
\usepackage{amsmath,amssymb}
\usepackage{amsthm}
\usepackage{nicefrac}       % compact symbols for 1/2, etc.
\usepackage{microtype}      % microtypography
\usepackage{cleveref}       % smart cross-referencing
\usepackage{graphicx}
\usepackage[numbers,sort&compress]{natbib}
\usepackage{doi}

\newtheorem{theorem}{Theorem}
\newtheorem{proposition}[theorem]{Proposition}
\newtheorem{conjecture}[theorem]{Conjecture}
\newtheorem{definition}[theorem]{Definition}

\title{Monotropic Artificial Intelligence: Toward a Cognitive Taxonomy of Domain-Specialized Language Models}

\newif\ifuniqueAffiliation
\uniqueAffiliationfalse

\ifuniqueAffiliation
\else
\usepackage{authblk}

\newbox{\orcid}\sbox{\orcid}{\includegraphics[scale=0.06]{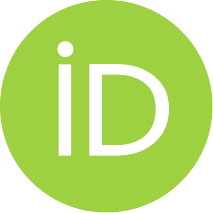}}
\author[1]{%
	Antonio de Sousa Leit\~{a}o Filho\thanks{\texttt{antonio@aiacontext.com}}%
}
\author[2]{%
	Allan Kardec Duailibe Barros Filho%
}
\author[1]{%
	Fabr\'{i}cio Saul Lima\thanks{\texttt{pdi@aiacontext.com}}%
}
\author[1]{%
	Selby Mykael Lima dos Santos\thanks{\texttt{legal@aiacontext.com}}%
}
\author[1]{%
	Rejani Bandeira Vieira Sousa\thanks{\texttt{data@aiacontext.com}}%
}
\affil[1]{Aia Context, S\~{a}o Lu\'{i}s, MA, Brazil}
\affil[2]{Federal University of Maranh\~{a}o, S\~{a}o Lu\'{i}s, MA, Brazil}
\fi

\renewcommand{\shorttitle}{Monotropic Artificial Intelligence}

\hypersetup{
pdftitle={Monotropic Artificial Intelligence: Toward a Cognitive Taxonomy of Domain-Specialized Language Models},
pdfsubject={cs.AI, cs.CL, cs.LG},
pdfauthor={Antonio de Sousa Leit\~ao Filho, Allan Kardec Duailibe Barros Filho, Fabr\'icio Saul Lima, Selby Mykael Lima dos Santos, Rejani Bandeira Vieira Sousa},
pdfkeywords={Monotropism, Domain Specialization, Language Models, Cognitive Architecture, AI Safety, Timoshenko Beam Theory},
}

\begin{document}
\maketitle

\begin{abstract}
The prevailing paradigm in artificial intelligence research equates progress with scale: larger models trained on broader datasets are presumed to yield superior capabilities. This assumption, while empirically productive for general-purpose applications, obscures a fundamental epistemological tension between breadth and depth of knowledge. We introduce the concept of \emph{Monotropic Artificial Intelligence}---language models that deliberately sacrifice generality to achieve extraordinary precision within narrowly circumscribed domains. Drawing on the cognitive theory of monotropism developed to understand autistic cognition, we argue that intense specialization represents not a limitation but an alternative cognitive architecture with distinct advantages for safety-critical applications. We formalize the defining characteristics of monotropic models, contrast them with conventional polytropic architectures, and demonstrate their viability through Mini-Enedina, a 37.5-million-parameter model that achieves near-perfect performance on Timoshenko beam analysis while remaining deliberately incompetent outside its domain. Our framework challenges the implicit assumption that artificial general intelligence constitutes the sole legitimate aspiration of AI research, proposing instead a cognitive ecology in which specialized and generalist systems coexist complementarily.
\end{abstract}


\section{Introduction}\label{sec:intro}

The history of artificial intelligence reveals a persistent fascination with generality. From the initial aspirations of the Dartmouth Conference \citep{mccarthy2006proposal} to contemporary large language models (LLMs), the field has implicitly treated capability breadth as the primary metric of progress \citep{chollet2019measure}. This orientation has produced remarkable achievements: models capable of conversing across diverse topics, generating creative content, and demonstrating what has been described as emergent reasoning abilities \citep{brown2020language, openai2023gpt4, wei2022emergent}.\footnote{The existence of genuinely ``emergent'' abilities in LLMs is debated. Schaeffer et al.\ \citep{schaeffer2023mirage} argue that such abilities are artifacts of the choice of nonlinear evaluation metrics.} Yet this paradigmatic commitment to generality contains an epistemological assumption that merits scrutiny---the presumption that cognitive architectures optimized for breadth represent the apex of intelligence.

Consider an alternative perspective drawn from human neurodiversity. The theory of monotropism, developed by Murray, Lesser, and Lawson to characterize autistic cognition, proposes that attention can be understood as a limited resource that individuals allocate differentially \citep{murray2005attention}. While neurotypical cognition tends toward ``polytropism''---distributing attention across multiple simultaneous interests---monotropic cognition channels attention intensively into restricted domains. This difference in attentional architecture produces characteristic strengths: monotropic individuals tend to develop deep engagement and detailed knowledge in their areas of interest, potentially achieving levels of proficiency that exceed what might be predicted by general measures of ability.

The parallel with artificial intelligence systems is striking. Large language models exemplify polytropic architecture: trained on web-scale corpora spanning countless domains, they acquire broad but depth-limited competence \citep{bender2020climbing}. This architecture supports impressive flexibility but introduces fundamental limitations. When applied to domains requiring precision---engineering calculations, medical diagnostics, legal reasoning---polytropic models exhibit characteristic failure modes: confident generation of plausible but incorrect outputs \citep{frieder2023mathematical, borji2023categorical}, difficulty in reliably calibrating the limits of their competence---although large models exhibit partial self-evaluation capability \citep{kadavath2022language}, this calibration remains insufficient for critical domains, and systematic difficulty with grounded numerical reasoning \citep{frieder2023mathematical}.

These limitations cannot be fully eliminated through additional scale, as they emerge from fundamental properties of autoregressive prediction \citep{kalai2024calibrated, xu2024hallucination}. A model optimized to predict the next token across all possible domains necessarily acquires statistical regularities that may diverge from domain-specific truths. The same training objective that enables creative generation also enables hallucination \citep{ji2023hallucination}; the same generality that permits flexible application also prevents guaranteed reliability. While scale reduces hallucination rates---as evidenced by improvement across successive model generations---recent theoretical results demonstrate that well-calibrated language models must necessarily hallucinate on long-tail topics \citep{kalai2024calibrated}.

This paper proposes an alternative architectural paradigm: \emph{Monotropic Artificial Intelligence}. We define monotropic models as language systems that deliberately restrict their domain of competence to achieve precision and reliability unattainable by polytropic approaches. Drawing on the cognitive theory of monotropism, we argue that this represents not a deficient cognitive architecture but a fundamentally \emph{different} one---one optimized for depth rather than breadth, certainty rather than flexibility, reliability rather than versatility.

Our contributions are threefold. First, we develop a formal taxonomy distinguishing monotropic from polytropic AI systems, identifying the defining characteristics of each architecture and the trade-offs they embody. Second, we articulate design principles for building monotropic models, grounded in physical validation, structured outputs, and intentional enforcement of boundaries. Third, we demonstrate the viability of this approach through Mini-Enedina, a 37.5-million-parameter model that achieves high structural reliability (perplexity 1.08; 100\% structural validity) on Timoshenko beam analysis while remaining deliberately incompetent outside its domain.

\section{Theoretical Foundations}\label{sec:theory}

\subsection{The Epistemology of Generalist AI}\label{subsec:epistemology}

The success of large language models has been attributed to scaling laws---empirical relationships demonstrating that model performance improves predictably with increases in parameters, data, and compute \citep{kaplan2020scaling, hoffmann2022training}. These observations have motivated an implicit research program: if scale produces improvement, then maximum scale produces maximum capability, and maximum capability approximates general intelligence.

This reasoning, while practically productive, conflates distinct epistemic categories. Performance on aggregate benchmarks does not imply reliable performance on any particular domain \citep{hendrycks2021measuring, liang2023helm}. A model achieving 90\% accuracy across diverse tasks may achieve only 60\% accuracy on specialized applications---and in safety-critical domains, the distinction between 90\% and 99.9\% reliability represents the difference between a useful tool and a dangerous liability.

More fundamentally, the scaling paradigm assumes a particular theory of knowledge: that expertise emerges from statistical regularities in sufficiently large datasets. This view treats knowledge as fundamentally \emph{correlational}---a model ``knows'' which word follows another because it has observed enough examples of the pattern. The distinction between correlational and causal knowledge has deep roots in the epistemology of science \citep{pearl2009causality} and in the symbol grounding problem \citep{harnad1990symbol}. Such correlational knowledge can approximate genuine understanding when the domain admits statistical regularities that align with fundamental truths, but diverges systematically when domains require \emph{grounded} knowledge derived from causal models of reality \citep{scholkopf2022causality}.

The distinction between correlational and grounded knowledge illuminates why polytropic models struggle with technical domains. Engineering calculations, for instance, must respect physical laws regardless of how frequently certain numerical patterns appear in training corpora. A model that has seen thousands of examples of ``2+2=4'' possesses correlational knowledge of this fact, but a model trained on physics-based simulations \citep{karniadakis2021physics, raissi2019physics} possesses grounded knowledge---its training data is guaranteed to be consistent with the physical laws that generated it, although the model itself still learns statistical patterns \emph{within} those validated data.

\subsection{Monotropism as Cognitive Architecture}\label{subsec:monotropism}

The theory of monotropism, articulated by Murray et al.\ \citep{murray2005attention} and recently connected to flow theory by Heasman et al.\ \citep{heasman2024flow}, offers a framework for understanding cognitive architectures that differ from the polytropic norm. Originally developed to characterize autistic cognition, monotropism posits that attention functions as a limited resource that can be allocated narrowly and intensely or broadly and superficially. Monotropic cognition channels attention into states we term attentional tunnels---intense foci in which a restricted domain receives concentrated processing resources.

This attentional architecture produces characteristic phenomena. Within their domains of focus, monotropic individuals tend to develop deep engagement and detailed knowledge that may exceed what is expected by general measures of ability. In extreme and relatively rare cases, monotropic attentional concentration can contribute to the savant phenomenon---exceptional abilities in specific domains such as mathematical calculation, musical performance, or visual art \citep{treffert2009savant}---although savant syndrome involves factors beyond monotropism and affects a minority of autistic individuals. The underlying principle, however, is robust: cognitive resources concentrated in narrow channels produce depth that distributed allocation can hardly match.

Importantly, monotropism does not represent cognitive deficiency but cognitive \emph{difference}---a central position in the neurodiversity paradigm \citep{singer1999neurodiversity, milton2012ontological}. The trade-offs it embodies---intense depth at the cost of reduced flexibility---may be disadvantageous in environments requiring constant context-switching but advantageous in environments rewarding sustained focus. Appropriate evaluation of monotropic cognition depends not on comparison with a polytropic norm but on assessment of the fit between cognitive architecture and environmental demands.

\begin{definition}[Monotropic Attentional Allocation]
Let $A$ be the total available attention and $D = \{d_1, d_2, \ldots, d_n\}$ the potential domains of focus. An attentional allocation function $f: D \rightarrow [0, A]$ is \emph{monotropic} if there exists a small subset $D^* \subset D$ with $|D^*| \ll |D|$ such that $\sum_{d \in D^*} f(d) \approx A$. In contrast, a \emph{polytropic} allocation distributes $A$ more uniformly across $D$.
\end{definition}

This formalization captures the essential characteristic of monotropism: the concentration of cognitive resources in restricted domains. Applied to artificial systems, it suggests that language models may adopt analogous architectural choices, concentrating their representational capacity on narrow domains rather than distributing it across general capabilities.

\subsection{From Human to Artificial Monotropism}\label{subsec:artificial}

The translation of monotropic principles from human cognition to artificial systems requires careful attention to both analogies and disanalogies. Human monotropism emerges from neural architecture and develops through experience; artificial monotropism must be deliberately engineered through training choices. Human monotropism admits degrees and variations; artificial monotropism can be precisely specified. Human monotropism interacts with emotion, motivation, and embodiment; artificial monotropism operates purely in the informational domain.

Despite these differences, the central insight transfers: cognitive architectures optimized for narrow domains can achieve capabilities not accessible, at comparable computational cost, to architectures optimized for breadth. Empirical precedents are instructive: AlphaFold~2 \citep{jumper2021alphafold} revolutionized protein structure prediction through radical specialization, and weather models such as GraphCast \citep{lam2023graphcast} surpassed traditional numerical systems via domain-specific architectures. Analogously, a language model trained exclusively on physics-based simulations of a particular engineering problem acquires knowledge qualitatively different from a model trained on web-scale corpora that happens to include some engineering content. The specialist model's knowledge is \emph{grounded} in the validated provenance of its training data; the generalist model's knowledge is \emph{interpolated} from statistical patterns that may or may not reflect physical reality.

\begin{proposition}[Monotropic Specialization Trade-off]
For a fixed model capacity $C$, domain coverage $D$, and domain-specific performance $P_d$, a monotropic architecture achieves $P_d^{mono} > P_d^{poly}$ for $d \in D^*$ (the focused domain) while accepting $P_d^{mono} < P_d^{poly}$ for $d \notin D^*$. Under the assumption that domains are sufficiently independent, the performance differential $\Delta P = P_d^{mono} - P_d^{poly}$ grows with the ratio $|D| / |D^*|$.
\end{proposition}

This proposition---whose formal validity rests on simplifying assumptions about domain independence and about the relationship between representational capacity and performance---formalizes the intuition, consistent with the No Free Lunch theorems \citep{wolpert1997nfl}, that specialization yields disproportionate benefits when the domain of specialization is narrow relative to the space of possible domains. A model that focuses its capacity on a single engineering problem while a competitor distributes capacity across thousands of topics tends to outperform on that domain by a margin that, empirically, appears to grow with the degree of specialization---although the precise scaling relationship remains an open question, depending on domain structure and cross-task transfer effects \citep{caruana1997multitask}.

\section{Formal Definition of Monotropic AI}\label{sec:definition}

Building on the theoretical foundations above, we now present a formal characterization of monotropic artificial intelligence. Our definition identifies four essential properties that distinguish monotropic from polytropic architectures.

\begin{definition}[Monotropic Language Model]
A language model $\mathcal{M}$ is \emph{monotropic} if and only if it satisfies the following four conditions:

\begin{enumerate}
    \item \textbf{Intentional Domain Restriction (IDR):} $\mathcal{M}$ is trained predominantly or exclusively on data from a well-defined domain $\mathcal{D}$, with deliberate measures to concentrate representational capacity on $\mathcal{D}$ \citep{gu2021pubmedbert, gururangan2020dapt}.

    \item \textbf{Depth over Breadth (DoB):} $\mathcal{M}$ is optimized for maximum performance within $\mathcal{D}$, accepting degraded or absent performance on $\mathcal{D}^c$.

    \item \textbf{Grounded Knowledge (GK):} Knowledge in $\mathcal{M}$ derives from training data generated by validated physical, mathematical, or logical models, ensuring that the learned statistical patterns reflect verified domain relationships \citep{karniadakis2021physics, willard2022integrating}.

    \item \textbf{Bounded Competence (BC):} $\mathcal{M}$ exhibits predictable and verifiable behavior within $\mathcal{D}$ and explicit incompetence outside $\mathcal{D}$.
\end{enumerate}
\end{definition}

Each condition merits elaboration. \textbf{Intentional Domain Restriction} distinguishes monotropic models from specialist models that emerge accidentally through biased data collection or underspecification of the training pipeline \citep{damour2022underspecification}. The restriction must be deliberate, reflecting an architectural choice rather than a data artifact. We note that IDR admits degrees: a model pre-trained on general data and subsequently fine-tuned exclusively on domain data represents a weaker form of IDR, while training from scratch on domain data (as in PubMedBERT \citep{gu2021pubmedbert} or Mini-Enedina) represents the strongest form.

\textbf{Depth over Breadth} captures the essential trade-off of the monotropic architecture. Where polytropic models seek to maximize aggregate performance across diverse domains, monotropic models maximize performance within their focused domain, explicitly accepting incompetence elsewhere. This is not a limitation to be overcome but a design feature to be embraced.

\textbf{Grounded Knowledge} distinguishes the epistemic character of monotropic expertise from the polytropic approximation. A monotropic model does not merely predict which tokens typically follow in discussions of its domain; it has learned to faithfully reproduce the input-output mappings generated by validated causal models. Whether this constitutes internalization of causal structure or precise interpolation of physics-generated data remains an open question in mechanistic interpretability research \citep{nanda2023grokking, li2023othello}. What is certain is that the provenance of the training data---generated by validated solvers rather than extracted from potentially erroneous text---enables reliability guarantees that correlational learning from unverified text corpora can hardly match.

\textbf{Bounded Competence} addresses the safety implications of specialization. A monotropic model should not merely fail outside its domain but fail \emph{gracefully and predictably}, employing mechanisms of selective prediction \citep{geifman2017selective} or out-of-distribution detection. Achieving this failure mode in practice requires explicit engineering of abstention mechanisms \citep{elyaniv2010selective}, an active area of research in competence-aware machine learning \citep{wen2025abstention}. The ideal failure mode is explicit recognition of incompetence rather than confident generation of plausible falsehoods.

Table~\ref{tab:comparison} synthesizes the distinctive characteristics of each architecture.

\begin{table}[!ht]
\caption{Comparative characteristics of polytropic and monotropic architectures.}\label{tab:comparison}
\centering
\begin{tabular}{lll}
\toprule
Characteristic & Polytropic (LLMs) & Monotropic \\
\midrule
Primary Objective & Generalization & Specialization \\
Training Data & Web-scale, diverse & Domain-specific, validated \\
Knowledge Type & Correlational & Grounded \\
Parameter Count & Typically billions & Minimum for target performance\textsuperscript{a} \\
Hallucination Risk & High & Low within domain \\
In-Domain Reliability & Variable & High within domain \\
Flexibility & High & Low \\
Failure Mode & Confident errors & Explicit incompetence \\
Safety Profile & Unpredictable & Bounded \\
\bottomrule
\end{tabular}
\vspace{0.5em}

{\footnotesize \textsuperscript{a}Parameter count is a consequence of domain scope, not a defining characteristic. Monotropic models may be larger for complex domains, just as smaller polytropic models exist (e.g., Phi-1, 1.3B parameters \citep{gunasekar2023textbooks}).}
\end{table}

\section{Epistemological Critique of the Scaling Paradigm}\label{sec:critique}

The monotropic framework enables a fundamental epistemological critique of the prevailing scaling paradigm. We argue that the assumption that larger and more general models necessarily represent progress contains hidden premises that fail in important application domains.

\subsection{The Conflation of Capability and Reliability}\label{subsec:conflation}

Scaling laws demonstrate that aggregate benchmark performance improves with model size. This observation, however, conflates two distinct properties: \emph{capability}---the range of tasks a model can attempt---and \emph{reliability}---the consistency with which a model succeeds at tasks within its capability range.

A 175-billion-parameter model \citep{brown2020language} can discuss Timoshenko beam theory, but its reliability on specific calculations remains uncertain \citep{frieder2023mathematical}. A 37.5-million-parameter model trained exclusively on Timoshenko analysis cannot discuss poetry or politics, but achieves high structural reliability (perplexity 1.08) within its domain. The question of which model is ``better'' has no context-free answer; it depends entirely on application requirements.

\begin{conjecture}[Capability-Reliability Trade-off]
For a fixed computational budget $B$, capability $C$ (measured as domain coverage) and reliability $R$ (measured as intra-domain accuracy) exhibit a trade-off relationship:
\begin{equation}
R \cdot C \leq f(B)
\end{equation}
where $f$ is an increasing function of the budget. Polytropic architectures maximize $C$; monotropic architectures maximize $R$ for a fixed $C^* \ll C_{max}$.
\end{conjecture}

This conjecture is consistent with the classical bias-variance trade-off \citep{belkin2019reconciling} and with the No Free Lunch theorems \citep{wolpert1997nfl}, although a formal proof remains an open problem.

This trade-off has been obscured by evaluation practices that emphasize aggregate metrics \citep{bowman2021benchmarking, ethayarajh2020utility}, a limitation that holistic approaches such as HELM \citep{liang2023helm} have begun to address. When models are evaluated by their average performance across diverse benchmarks, architectures that maximize capability appear superior. But when evaluation focuses on worst-case reliability within specific domains---the relevant metric for safety-critical applications---the advantages of specialization emerge.

\subsection{The Impossibility of Universal Grounding}\label{subsec:impossibility}

A deeper critique concerns the epistemic limits of correlational learning. Polytropic models acquire knowledge by observing statistical regularities in text. This learning mechanism can approximate genuine understanding when textual patterns reliably indicate underlying truths, but fails when patterns and truths diverge.

Consider a model trained on internet text about physics. The training corpus contains both correct physics (textbooks, articles) and incorrect physics (misconceptions, science fiction, jokes). The model learns to predict tokens that \emph{typically} follow in discussions of physics, but ``typical'' patterns may diverge from correct patterns. The model cannot distinguish authoritative from erroneous sources purely from statistical regularities; it can only learn that certain patterns are more frequent.

This limitation is not fully remediable through scale \citep{kalai2024calibrated}. A larger model trained on more data acquires more precise estimates of statistical regularities, but more precise estimates of potentially erroneous patterns remain erroneous. The model becomes more confident in its correlational knowledge without becoming more grounded in causal truth. We note that this argument applies specifically to the standard pretraining paradigm; architectural modifications such as external tool use or hybrid neuro-symbolic systems may address these limitations by means other than scale.

The monotropic architecture addresses this limitation by restricting training data to validated sources. A model trained exclusively on physics simulations that have been verified against analytical solutions acquires knowledge whose training signal is grounded in validated physics, substantially reducing---though not eliminating---the risk of physically implausible outputs. It may ``know'' less than a polytropic model, but what it knows is aligned with physical reality within the bounds of its training distribution.

\subsection{The Safety Implications of Unbounded Competence}\label{subsec:safety}

Polytropic models exhibit a characteristic we term \emph{unbounded competence}---the willingness to generate responses on any topic regardless of the model's actual expertise. This phenomenon is related to the overconfidence and sycophancy documented in the literature \citep{kadavath2022language, sharma2023sycophancy}. In base models, this property emerges directly from the training objective: models are optimized to predict likely continuations, and refusing to answer is rarely the most likely continuation. While post-training alignment (RLHF) introduces refusal on safety-sensitive queries \citep{ouyang2022instructgpt}, models do not systematically learn to refuse on \emph{epistemic} grounds---readily generating responses on topics where their training data does not provide sufficient grounding for reliable answers.

Unbounded competence creates safety risks in proportion to user confidence. A user who correctly calibrates uncertainty---treating all model outputs as potentially erroneous---can extract value from polytropic models while avoiding harm. But the literature on automation bias demonstrates that users frequently miscalibrate, especially when automated systems present outputs with high confidence \citep{parasuraman2010complacency}. Studies specific to LLMs confirm this tendency toward user overconfidence \citep{vasconcelos2023overreliance}. The failure mode is pernicious: the model generates plausible, confident, incorrect outputs, and the user lacks the expertise to recognize the error.

The monotropic architecture inverts this failure mode. A model with bounded competence refuses to answer outside its domain, transforming potential confident errors into explicit acknowledgments of incompetence. The user may be frustrated by the model's limitations, but the risk of being misled by confident falsehoods is substantially reduced compared to polytropic alternatives.

\begin{proposition}[Safety Advantage of Bounded Competence]
Following the classical framework of probabilistic risk analysis \citep{kaplan1981risk}, let $H_d$ be the harm from an incorrect response in domain $d$ and $\epsilon_d$ the error rate of model $\mathcal{M}$ in that domain. The expected harm from using $\mathcal{M}$ is:
\begin{equation}
\mathbb{E}[H] = \sum_{d \in D} P(q \in d) \cdot \epsilon_d \cdot H_d
\end{equation}
where $\mathbb{E}[H]$ denotes the expected value of harm, $D$ is the set of all domains, $P(q \in d)$ is the probability of a query $q$ belonging to domain $d$, $\epsilon_d$ is the error rate in domain $d$, and $H_d$ is the harm associated with errors in that domain. For a monotropic model with $\epsilon_d \approx 0$ for $d \in D^*$ and explicit refusal for $d \notin D^*$, expected harm is minimized when queries are restricted to $D^*$. This result follows directly from the non-negativity of all terms in the sum. While mathematically elementary, this formalization makes explicit the intuition that bounded competence reduces expected harm by eliminating contributions from domains where the model is unreliable.
\end{proposition}

\section{Design Principles for Monotropic Systems}\label{sec:principles}

The theoretical framework above yields practical design principles for building monotropic AI systems. We identify four core principles that distinguish monotropic engineering from conventional LLM development.

\subsection{The Principle of Intentional Exclusion}\label{subsec:exclusion}

\begin{quote}
\emph{A monotropic model should be designed to exclude, not merely to include.}
\end{quote}

Conventional LLM development has historically sought to maximize data inclusion: more data, more diverse data, more capabilities---and compute-optimal scaling results suggest that prior models were in fact undertrained on data relative to their size \citep{hoffmann2022training}. Monotropic development inverts this orientation. The critical engineering decisions concern what to \emph{exclude} from training data, which queries to \emph{refuse}, which outputs to \emph{prevent}.

This principle has concrete implications. Training data must be curated not merely for quality but for domain restriction. Out-of-domain queries must be detected and rejected rather than speculatively answered \citep{wen2025abstention}. The model must learn not only what to say but what \emph{not} to say.

\subsection{The Principle of Physical Grounding}\label{subsec:grounding}

\begin{quote}
\emph{Knowledge in a monotropic model should derive from validated sources with traceable data provenance, not from unverified text corpora.}
\end{quote}

Monotropic training data must derive from validated sources---including physical simulations, curated experimental measurements, expert-verified knowledge bases, or formal derivations---rather than unstructured text corpora \citep{karniadakis2021physics}. Each training example should have clear data provenance: traceability to a validated computational model, verified measurement, or formal derivation. The statistical structure of the training data should be determined by valid domain relationships, ensuring that the patterns learned by the model faithfully reflect the causal structure of the data-generating process.

Mini-Enedina exemplifies this principle. Training data consists of 60,000 samples generated by solving Timoshenko beam equations numerically. Each sample pairs a problem specification with its validated solution. The model learns to reproduce the mapping from specifications to solutions, but this mapping is grounded in physics: the training data contains numerically validated solutions because the data-generation process imposes correctness criteria at multiple levels, including comparison with analytical benchmarks where available. Only samples that pass all verification levels enter the training set.

\subsection{The Principle of Structured Verification}\label{subsec:verification}

\begin{quote}
\emph{Monotropic outputs should be structured to facilitate verification by humans or automated systems.}
\end{quote}

The value of a monotropic model derives from its reliability, and reliability must be verifiable. This requires outputs in structured formats that separate reasoning from conclusions, expose intermediate calculations, and enable systematic verification---connecting to research on step-by-step verification \citep{lightman2023verify} and transparency through chain-of-thought reasoning \citep{wei2022chain}.

The Harmony-Enedina format implements this principle. Each response contains distinct channels for analysis, reasoning, code, and verification. The code channel contains executable implementations that can be independently validated. The verification channel reports multi-level checks confirming solution correctness. Users can verify not merely that the model produced an answer but \emph{how} the model arrived at that answer.

\subsection{The Principle of Minimal Size}\label{subsec:minimal}

\begin{quote}
\emph{A monotropic model should be the smallest model that achieves target performance within its domain.}
\end{quote}

The scaling paradigm assumes that larger models are better models. Monotropic design inverts this assumption, echoing the Minimum Description Length (MDL) principle from information theory \citep{rissanen1978mdl}: among models that achieve target performance, prefer the most parsimonious. Smaller models require fewer computational resources for training and inference \citep{strubell2019energy, patterson2021carbon}, enable deployment on edge devices, and reduce environmental impact.

Mini-Enedina's 37.5 million parameters represent several orders of magnitude fewer than contemporary LLMs, which typically range from tens of billions to over one trillion parameters. Yet within its domain, it achieves near-perfect performance---a level of reliability that polytropic models have not demonstrated for grounded numerical engineering calculations \citep{frieder2023mathematical}. This is not despite but \emph{because of} its size: the model's capacity is concentrated on a narrow domain rather than distributed across countless topics.

\section{Case Study: Mini-Enedina}\label{sec:case}

We validate the monotropic framework through Mini-Enedina, a language model designed for structural analysis of power transmission shafts using Timoshenko beam theory.

\subsection{Domain Specification}\label{subsec:domain}

The domain $\mathcal{D}$ encompasses analysis of cylindrical shafts under combined loading, governed by Timoshenko beam equations that account for shear deformation \citep{timoshenko1921correction}. The analysis hierarchy comprises three levels of increasing complexity:

\begin{enumerate}
    \item \textbf{Bachelor Level:} Deflection analysis---shear force $V(x)$, bending moment $M(x)$, slope $\theta(x)$, deflection $w(x)$
    \item \textbf{Master Level:} Stress analysis---von Mises equivalent stress $\sigma_{VM}$, yield safety factor
    \item \textbf{Doctor Level:} Fatigue analysis---Marin factors, modified endurance limit $S_e$, Goodman criterion \citep{shigley2011mechanical}
\end{enumerate}

This domain is narrow by design. Mini-Enedina cannot analyze plates, shells, or other structural elements. It cannot perform finite element analysis or handle nonlinear materials. These restrictions are features, not defects: they enable the concentration of the model's capacity on a well-defined class of problems.

\subsection{Training Data Generation}\label{subsec:training}

Training data consists of 60,000 synthetic samples generated by the Factorium framework \citep{factorium2026}. Each sample comprises:

\begin{itemize}
    \item A problem specification in natural language (Portuguese)
    \item Complete Python solver code implementing Timoshenko analysis
    \item Numerical results validated against analytical solutions where available
    \item Multi-level verification confirming physical plausibility
\end{itemize}

The data-generation process enforces physical grounding. Parameters are sampled from physically realistic distributions. Solutions are computed using validated numerical methods. Results are verified against conservation laws and analytical benchmarks. Only samples that pass all verification levels enter the training set. Table~\ref{tab:dataset} presents the statistics of the resulting dataset.

\begin{table}[!ht]
\caption{Training dataset statistics for Mini-Enedina.}\label{tab:dataset}
\centering
\begin{tabular}{lrr}
\toprule
Level & Samples & Tokens \\
\midrule
Bachelor & 20,000 & 168M \\
Master & 20,000 & 198M \\
Doctor & 20,000 & 255M \\
\midrule
\textbf{Total} & \textbf{60,000} & \textbf{621M} \\
\bottomrule
\end{tabular}
\end{table}

\subsection{Model Architecture}\label{subsec:architecture}

Mini-Enedina employs a dense transformer architecture optimized for the Timoshenko domain, namely:

\begin{itemize}
    \item Parameters: 37.5M
    \item Layers: 7
    \item Attention heads: 8
    \item Model dimension: 512
    \item Feed-forward dimension: 2,048
    \item Vocabulary: 8,012 tokens (8,000 BPE + 12 Harmony-Enedina tokens)
    \item Maximum sequence length: 14,336 tokens
\end{itemize}

Training employed multidimensional curriculum learning \citep{bengio2009curriculum} that simultaneously varies the mixing ratio between levels and the difficulty thresholds per level. Crucially, all three levels are introduced from the initial phase, differentiated only by the difficulty thresholds controlling which samples from each level are eligible. The Foundation phase includes 70\% Bachelor, 20\% Master, and 10\% Doctor; subsequent phases progressively increase the proportion of more complex levels until reaching uniform distribution in the final phase. This early exposure to all levels, including Doctor samples with long sequences, ensures that the model learns completion patterns from the outset, avoiding the sequential training regime that would induce catastrophic forgetting \citep{mccloskey1989catastrophic, kirkpatrick2017overcoming}.

\subsection{Results}\label{subsec:results}

Evaluation on a held-out test set of 6,000 samples demonstrates the viability of the monotropic architecture. Table~\ref{tab:results} presents performance metrics by complexity level.

\begin{table}[!ht]
\caption{Mini-Enedina performance on the test set.}\label{tab:results}
\centering
\begin{tabular}{lrrr}
\toprule
Metric & Bachelor & Master & Doctor \\
\midrule
Test Loss & 0.0733 & 0.0804 & 0.0825 \\
Perplexity & 1.08 & 1.08 & 1.09 \\
Structural Validity & 100\% & 100\% & 100\% \\
Numerical Grounding & 100\% & 100\% & 100\% \\
Correct Stop Token & 97\% & 100\% & 85\% \\
\bottomrule
\end{tabular}
\end{table}

These results merit careful interpretation. A perplexity of 1.08 indicates that the model achieves near-deterministic prediction of validation sequences. It is important to note that part of this low perplexity reflects the deterministic structure of the outputs---Python boilerplate code, structural Harmony tokens, and LaTeX formatting are highly predictable---such that the effective perplexity over informative tokens (numerical values, physical reasoning) is likely higher. Nonetheless, this could be problematic for a polytropic model, suggesting overfitting, but represents success for a monotropic model: the model has learned to replicate with high fidelity the input-output mappings generated by the validated Timoshenko beam analysis solvers. Note that the correct stop token rate, while high (97\% Bachelor, 100\% Master), drops to 85\% at the Doctor level, indicating that longer sequences ($\sim$14K tokens) represent a challenge for the current architecture.

\subsection{Demonstration of Bounded Competence}\label{subsec:bounded}

To verify bounded competence, we tested Mini-Enedina with out-of-domain queries. When presented with questions about topics outside Timoshenko beam analysis---historical events, literary interpretation, general physics---the model exhibits degraded performance characterized by repetitive token generation \citep{holtzman2020curious} and failure to produce coherent responses. This failure mode, while not constituting graceful degradation in the strict engineering sense (which would imply explicit refusal), is \emph{detectable}: a user immediately identifies the output as non-functional. This behavior contrasts with the tendency of polytropic models toward confident hallucination \citep{ji2023hallucination}, where incorrect outputs are superficially indistinguishable from correct ones.

\section{Philosophical Implications}\label{sec:philosophy}

The monotropic framework carries implications beyond engineering practice, challenging assumptions embedded in contemporary AI research and philosophy of mind.

\subsection{Against the AGI Presumption}\label{subsec:agi}

The prevailing narrative of AI progress positions artificial general intelligence as the ultimate aspiration \citep{mccarthy2006proposal, bostrom2014superintelligence}. From this perspective, specialized systems represent waypoints on the path to generality---useful but inherently limited approximations to the true goal.

Monotropic AI challenges this narrative. If specialization enables, at comparable computational cost, capabilities that generalist systems cannot practically replicate---and verifiability guarantees that generalist training cannot provide regardless of scale---then AGI may not represent the apex of artificial intelligence but merely one point in a space of possible cognitive architectures. Just as human cognitive diversity includes both polytropic and monotropic styles, artificial intelligence may encompass a cognitive ecology of specialized and generalist systems serving complementary roles.

This perspective has ethical implications. The drive toward AGI has been justified partly by expected benefits---systems capable of solving arbitrary problems could address humanity's greatest challenges. But monotropic systems may offer safer paths to many of those benefits \citep{russell2019human}: bounded, verifiable, auditable systems that solve specific problems without the risks associated with unbounded artificial agency.

\subsection{Lessons from Neurodiversity}\label{subsec:neurodiversity}

The translation of monotropism from human cognition to artificial systems invites reflection on the relationship between natural and artificial intelligence. Human monotropism is not chosen but experienced; it emerges from neural architecture shaped by genetics and development \citep{murray2005attention}. Artificial monotropism is deliberately engineered, a design choice rather than an innate characteristic.

Yet the functional parallels are illuminating. In both cases, specialized architecture enables capabilities that generalist architecture can hardly match at comparable computational cost. In both cases, specialization trades flexibility for depth. In both cases, appropriate evaluation requires considering the fit between architecture and application rather than comparing against a presumed general norm.

This parallel suggests that insights from neurodiversity research may inform AI design, and conversely that AI research may illuminate aspects of human cognitive diversity. Speculatively, Monotropic AI may serve not only as an application inspired by autistic cognition but as a functional analogy---not a cognitive model in the strict sense---for investigating the trade-offs that concentrated attentional allocation embodies. This hypothesis remains at the level of conceptual inspiration and requires future empirical validation.

\subsection{The Epistemology of Bounded Systems}\label{subsec:bounded_epistemology}

Monotropic AI embodies a distinctive epistemological stance: the value of knowing what one does not know. Polytropic systems exhibit what we term the ``curse of competence''---an epistemic extension of automation bias \citep{parasuraman2010complacency}---where the ability to generate plausible responses on any topic creates the illusion of expertise where none exists. Unlike classical automation bias, which concerns specific decisions, the curse of competence makes confidence calibration extremely difficult in practice: users cannot easily distinguish authoritative from speculative outputs without independent verification.

Monotropic systems invert this dynamic. Their bounded competence is epistemically transparent: users know the domain in which the system is reliable and can calibrate trust appropriately. The system's explicit incompetence outside its domain reduces the risk of excessive reliance and encourages appropriate human oversight.

This epistemological transparency has implications for human-AI collaboration. When humans can accurately model the capabilities and limitations of an AI system, they can use it more effectively and avoid the pitfalls of miscalibrated trust \citep{vasconcelos2023overreliance}. Bounded systems may thus enable more productive collaboration than unbounded systems, despite---or because of---their narrower capabilities.

\section{Implications for AI Safety and Design}\label{sec:implications}

The monotropic framework has direct implications for the design of safe and reliable AI systems, particularly in domains where errors carry significant consequences.

\subsection{Safety-Critical Applications}\label{subsec:safety_critical}

Engineering, medicine, law, and finance involve decisions where incorrect AI outputs could cause substantial harm \citep{amodei2016concrete}. The unbounded competence of polytropic models makes them poorly suited for these domains: they tend to generate outputs regardless of their actual reliability, and users cannot easily distinguish reliable from unreliable responses.

Monotropic architecture offers a safer alternative for specific applications within these domains. A monotropic model for structural analysis, trained on validated physics simulations, provides levels of reliability and traceability that polytropic models can hardly match at comparable cost. The trade-off---inability to handle queries outside the training domain---constitutes a deliberate and desirable limitation: it substantially reduces the risk of the model generating confidently incorrect outputs on unfamiliar problems.

\subsection{Complementary Architectures}\label{subsec:complementary}

Monotropic and polytropic systems need not compete; they can complement each other in hybrid architectures. Drawing on the mixture-of-experts principle \citep{jacobs1991adaptive, shazeer2017outrageously}---though with independently trained specialists rather than jointly optimized ones---a polytropic system can serve as a ``router,'' identifying which specialist system should handle a particular query. Monotropic systems would provide reliable responses within their domains, while the polytropic system would handle genuinely general queries that do not require domain-specific expertise.

This architecture preserves the flexibility of polytropic systems while enabling the reliability of monotropic systems where it matters most. Users benefit from both breadth and depth without compromising either.

\subsection{Auditability and Accountability}\label{subsec:auditability}

The structured outputs of monotropic systems such as Mini-Enedina facilitate auditing and accountability. When a system produces a response through explicit reasoning steps and verifiable code, reviewers can trace the logic that produced each conclusion. This transparency enables meaningful human oversight: although verification of physical correctness requires domain expertise, the reasoning structure and code executability can be independently audited.

Polytropic systems resist such auditing. Their reasoning is implicit in billions of parameters; their outputs emerge from statistical patterns that, despite advances in mechanistic interpretability \citep{nanda2023grokking}, remain largely opaque to human inspection. Even when polytropic systems produce chain-of-thought explanations, these explanations may not faithfully reflect the computations that actually produced the output \citep{turpin2023unfaithful}, and preliminary evidence suggests that this unfaithfulness may increase with model scale \citep{lanham2023measuring}.

\section{Limitations and Future Directions}\label{sec:limitations}

The monotropic framework has limitations that merit acknowledgment, and our current implementation leaves substantial room for future development.

\subsection{Domain Boundary Definition}\label{subsec:boundary}

The monotropic framework presupposes that domains can be clearly delineated, but domain boundaries are often fuzzy in practice. Where does Timoshenko beam analysis end and general structural mechanics begin? How should a monotropic system handle queries that partially overlap its domain?

These questions connect to the established literature on out-of-distribution (OOD) detection \citep{hendrycks2017baseline, yang2024generalized}. Existing OOD methods detect when inputs lie outside the training distribution using confidence scores, energy-based measures, or density estimation. The monotropic architecture provides a structural analog: by restricting training exclusively to the target domain, the learned representations naturally diverge from out-of-domain inputs. However, implementing probabilistic domain membership---a formal decision boundary for the monotropic model---remains an open engineering problem. Other approaches include hierarchical domain structures and explicit uncertainty quantification for borderline queries.

\subsection{Knowledge Transfer}\label{subsec:transfer}

A limitation of extreme specialization is the inability to leverage knowledge across related domains. A polytropic model that learns about beams may transfer some of that knowledge to plates; a monotropic model trained exclusively on beams cannot.

Future work may explore ``monotropic ensembles''---collections of specialized models with mechanisms for selective knowledge sharing. Such architectures may preserve monotropic reliability while enabling limited transfer across related domains.

\subsection{Scaling Monotropic Systems}\label{subsec:scaling}

Our current implementation validates the monotropic concept at modest scale (37.5M parameters, single domain). Questions remain about how monotropic principles apply to larger models and broader---though still restricted---domain collections.

We hypothesize that monotropic advantages persist at larger scales: a 1B-parameter model trained on structural mechanics would outperform a general 100B-parameter model on structural problems. Indirect evidence is encouraging---small models trained on curated synthetic data have already demonstrated performance comparable to models orders of magnitude larger on specific tasks \citep{gunasekar2023textbooks}---but direct empirical validation for the monotropic case remains needed.

\subsection{Evaluation Scope}\label{subsec:evaluation_scope}

The current evaluation presents two methodological limitations that merit explicit acknowledgment.

First, we did not include a comparison with a generalist language model fine-tuned on the same Timoshenko dataset. Such a baseline---for example, a model of comparable parameter count pre-trained on general text and subsequently fine-tuned on the 60,000 synthetic samples---would isolate the contribution of architectural specialization from the contribution of domain data. It remains an open empirical question whether training exclusively on domain data confers advantages beyond those achievable via fine-tuning a generalist foundation model. This comparison constitutes a priority for future work.

Second, the evaluation relies exclusively on automated metrics applied to a held-out partition from the same synthetic training distribution. While structural validity and numerical grounding are verified algorithmically, no domain expert assessment was conducted to verify whether Mini-Enedina's outputs meet the professional standards expected of Timoshenko shaft analysis in engineering practice. Synthetic benchmarks may overestimate real-world utility if natural-language formulations by practicing engineers systematically diverge from the template-generated queries used in training. Expert evaluation protocols should be included in future work to validate generalization beyond the synthetic distribution.

\section{Conclusion}\label{sec:conclusion}

We have introduced Monotropic Artificial Intelligence as a cognitive taxonomy for domain-specialized language models. Drawing on the theory of monotropism from cognitive science, we argue that intense specialization represents not a limitation but an alternative cognitive architecture with distinct advantages for applications requiring reliability over versatility.

Our formal definition identifies four essential properties of monotropic systems: intentional domain restriction, depth over breadth, grounded knowledge, and bounded competence. These properties distinguish monotropic from polytropic architectures and yield practical design principles for engineering specialized AI systems.

The Mini-Enedina case study demonstrates that monotropic architecture can be both viable and effective. With several orders of magnitude fewer parameters than contemporary LLMs, Mini-Enedina achieves high fidelity in reproducing Timoshenko beam analyses while exhibiting appropriate incompetence outside its domain. This result challenges the assumption that AI progress requires ever-larger and more general models.

The implications extend beyond engineering practice to the philosophy of AI. If specialized systems can achieve, at comparable cost, capabilities and verifiability guarantees that generalist systems can hardly replicate, then the aspiration toward artificial general intelligence may represent one architectural choice among many rather than the singular goal of intelligent systems research. Just as human cognitive diversity encompasses both polytropic and monotropic styles, artificial intelligence may encompass a cognitive ecology in which specialized and generalist systems can coexist complementarily.

We call for expanded research into monotropic architectures for safety-critical domains. Where reliability matters more than flexibility, where errors carry significant consequences, and where domain boundaries can be clearly specified, monotropic systems may offer safer and more reliable alternatives to polytropic approaches. The future of AI may lie not in ever-larger general models but in a diverse ecosystem of specialized systems, each achieving excellence within its domain while recognizing its limitations beyond.

\section*{Acknowledgments}

This work was supported by Aia Context Ltda.\ and by FINEP---Financiadora de Estudos e Projetos (\url{http://www.finep.gov.br/}), a Brazilian government agency for science, technology, and innovation linked to the Ministry of Science, Technology, and Innovation (MCTI), under Contract No.\ 03.25.0080.00. The authors thank the research community working on neurodiversity and cognitive science for developing the theoretical foundations that inspired this work.

\section*{Declarations}

\begin{itemize}
\item \textbf{Conflict of interest:} The authors declare no competing financial interests or known personal relationships that could have influenced the work reported in this article.
\item \textbf{Data availability:} The trained model and weights are publicly available at \url{https://huggingface.co/aiacontext/mini-enedina}. The training dataset is available at \url{https://huggingface.co/datasets/aiacontext/mini-enedina-dataset}.
\end{itemize}

%%% Use thebibliography for inline references

\end{document}